\documentclass[11pt,a4paper]{article}
\usepackage[utf8]{inputenc}
\usepackage{amsmath}
\usepackage{amsfonts}
\usepackage{amssymb}
\usepackage{graphicx}
\usepackage{hyperref}
\usepackage{xcolor}
\usepackage{geometry}
\usepackage{booktabs} 
\usepackage{algorithm}
\usepackage{algpseudocode}
\usepackage{multirow} 

\hypersetup{
    colorlinks=true,
    linkcolor=blue,
    filecolor=magenta,      
    urlcolor=cyan,
    pdftitle={HumorPlanSearch: Structured Planning and HuCoT for Contextual AI Humor},
    pdfpagemode=FullScreen,
}

\title{\textbf{HumorPlanSearch: Structured Planning and HuCoT for Contextual AI Humor}}
\author{
  Shivam Dubey \\
  \small{Indian Institute of Technology Madras} \\
  \small{\texttt{23f1002279@ds.study.iitm.ac.in}}
}
\date{\today}

\begin{document}

\maketitle

\begin{abstract}
Automated humor generation with Large Language Models (LLMs) often yields jokes that feel generic, repetitive, or tone-deaf because humor is deeply situated—it hinges on the listener’s cultural background, mindset, and immediate context. We introduce \textbf{HumorPlanSearch}, a modular pipeline that explicitly models context through: (1) \textbf{Plan-Search} for diverse, topic-tailored strategies; (2) \textbf{Humor Chain-of-Thought (HuCoT)} templates capturing cultural and stylistic reasoning; (3) a \textbf{Knowledge Graph} to retrieve and adapt high-performing historical strategies; (4) \textbf{novelty filtering} via semantic embeddings; and (5) an iterative \textbf{judge-driven revision} loop. To evaluate context sensitivity and comedic quality, we propose the \textbf{Humor Generation Score (HGS)}, which fuses direct ratings, multi-persona feedback, pairwise win-rates, and topic relevance. In experiments across nine topics with feedback from 13 human judges, our full pipeline (KG + Revision) boosts mean HGS by 15.4\% (p < 0.05) over a strong baseline. By foregrounding context at every stage—from strategy planning to multi-signal evaluation—HumorPlanSearch advances AI‐driven humor toward more coherent, adaptive, and culturally attuned comedy.
\end{abstract}

\section{Introduction}

The automated generation of humor remains a significant challenge in artificial intelligence, requiring not just language fluency but also creativity, contextual awareness, and the ability to handle subjective reasoning. While recent work has explored culturally-specific humor generation [Zhong et al., 2023], many systems still fall short.

Standard "prompt and sample" techniques with Large Language Models (LLMs), for instance, often produce generic, templated, or repetitive jokes. Furthermore, LLM-based evaluations of humor can be inconsistent and fail to capture the multifaceted nature of comedy.

To address these shortcomings, we introduce \textbf{HumorPlanSearch}, a comprehensive pipeline designed to systematically improve the quality, novelty, and contextual relevance of AI-generated humor. Our main contributions are as follows:
\begin{itemize}
    \item We introduce a novel pipeline, \textbf{HumorPlanSearch}, that integrates strategic planning, a knowledge graph (KG), and iterative revision.
    \item We propose a multi-faceted evaluation metric, the \textbf{Humor Generation Score (HGS)}, for more robust automated assessment.
    \item We demonstrate the effectiveness of culturally-aware \textbf{Humor Chain-of-Thought (HuCoT)} templates for generating nuanced humor.
\end{itemize}

This holistic approach is designed to foster more sophisticated, adaptive, and genuinely amusing AI-generated content.

\section{Methodology: The HumorPlanSearch Pipeline}

The HumorPlanSearch pipeline is a multi-stage process that begins with an input topic and produces a ranked list of jokes with their reasoning traces.

\begin{figure}[h!]
    \centering
    \includegraphics[width=0.6\textwidth]{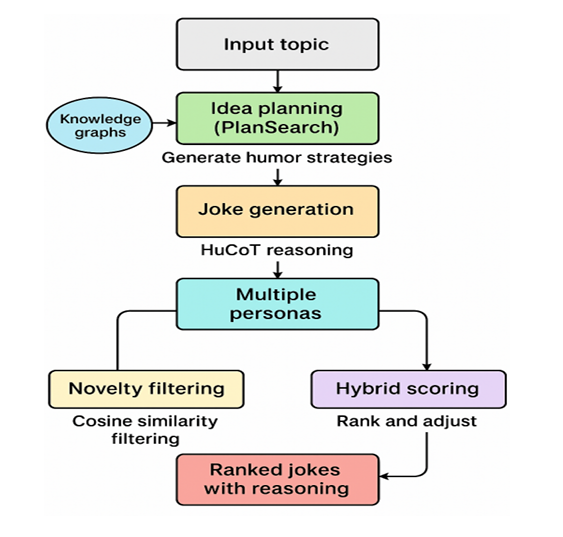} 
    \caption{The HumorPlanSearch Pipeline Flowchart.}
    \label{fig:pipeline}
\end{figure}

\subsection{Plan Generation}
The process starts with the \textbf{Strategy LLM} (default: \texttt{gemma2-9b-it}) generating diverse humor approaches. The choice of $N=12$ initial strategies and $max\_k=3$ hybrid combinations was determined empirically to balance strategic diversity with computational cost.
\begin{itemize}
    \item \textbf{First-Order Observations}: Initially, $N=12$ diverse humor approaches are generated.
    \item \textbf{Second-Order Hybrids}: Pairs of the initial observations are then combined to create more complex, hybrid strategies, up to a maximum of $k \times 3$ (where $k=3$).
    \item \textbf{Knowledge Graph (KG) Enhancement}: High-performing historical strategies are retrieved from a persistent NetworkX knowledge graph and adapted to the current topic, enriching the pool of potential strategies.
\end{itemize}

\subsection{Humor Chain-of-Thought (HuCoT) Generation}
We implement three distinct style variants for joke generation, each with a specific multi-step reasoning process executed by the \textbf{Joke Generation LLM} (default: \texttt{llama-3.3-70b-versatile}). Table \ref{tab:hucot} summarizes the templates.

\begin{table}[h!]
\centering
\caption{Humor Chain-of-Thought (HuCoT) Style Templates.}
\label{tab:hucot}
\begin{tabular}{@{}lcl@{}}
\toprule
\textbf{Style} & \textbf{Steps} & \textbf{Key Reasoning Focus} \\ \midrule
Generic & 8 & Universal comedic structures, timing, and punchline design. \\
Indian & 6 & Cultural micro-context, shared experiences, and delivery. \\
Gen Z-Indian & 7 & Dark reality, meme culture, and cultural code-switching. \\ \bottomrule
\end{tabular}
\end{table}

\subsection{Novelty Filtering}
To ensure a diverse output, we filter out jokes that are semantically similar. In our experiments, the 0.75 threshold removed an average of 18\% of generated jokes.
\begin{itemize}
    \item \textbf{Algorithm}: Each new joke is compared to the set of jokes already kept, using cosine similarity from sentence embeddings (\texttt{all-MiniLM-L6-v2}).
    \item \textbf{Threshold}: Jokes with a similarity score $\ge 0.75$ are removed.
\end{itemize}

\subsection{Hybrid Scoring}
We use a weighted combination of four complementary signals to evaluate the generated jokes, with evaluations performed by a \textbf{Judge LLM} (default: \texttt{llama3-70b-8192}). The final HGS is calculated as follows:
\begin{equation}
\text{HGS} = \sum_{i=1}^{4} w_i \times s_i, \quad \sum w_i=1
\end{equation}
where $w_i$ is the weight and $s_i$ is the score for each signal. The four signals are:
\begin{enumerate}
    \item \textbf{Direct Vote Scoring}: A direct 1-5 rating of the joke.
    \item \textbf{Multi-Persona Scoring}: Three distinct personas (\textbf{Enthusiastic Fan, Critical Critic, Academic Analyst}) evaluate both the joke and its HuCoT reasoning.
    \item \textbf{Pairwise Win-Rate}: A binary comparison against other jokes to determine a win rate.
    \item \textbf{Topic Relevance}: The cosine similarity between the joke and the initial topic prompt.
\end{enumerate}

\subsection{Judge-Guided Plan Revision}
The pipeline includes an iterative refinement loop (default: max 2 iterations) to improve humor strategies. A strategy is revised if its average score falls below 6.0 and the suggested revision is projected to improve the score by at least 0.2 (the \textbf{improvement threshold}). Algorithm \ref{alg:revision} summarizes the loop.

\begin{algorithm}
\caption{Judge-Guided Plan Revision Loop}
\label{alg:revision}
\begin{algorithmic}[1]
\State \textbf{Input:} Current strategies $S$, Style, Iterations $I_{max}$
\For{$i = 1$ to $I_{max}$}
    \State Generate jokes $J$ from strategies $S$
    \State Filter $J$ for novelty $\rightarrow J_{filtered}$
    \State Score $J_{filtered}$ using multi-persona evaluation
    \State Calculate per-strategy performance $P_s$ for each $s \in S$
    \State Get improvement suggestions $U_s$ from Judge LLM
    \State $S_{low} \gets \{s \in S \mid P_s < 6.0\}$
    \State $S_{high} \gets \{s \in S \mid P_s \ge 7.0\}$
    \State Generate revised strategies $S'_{low}$ from $S_{low}$ and $U_s$
    \State $S \gets S'_{low} \cup S_{high}$
\EndFor
\State \textbf{Return:} Refined strategies $S$
\end{algorithmic}
\end{algorithm}

\section{Evaluation and Findings}

\subsection{Preliminary Human Evaluation}
An initial study with 13 human judges from mixed cultural backgrounds (ages 20-45) revealed that our generic HuCoT could be perceived as "dry." This insight directly led to the development of the multi-style HuCoT approach and a multi-model architecture. The study also showed that smaller LLMs (e.g., Llama3-8B class) were less effective for nuanced joke generation.

\subsection{Configuration and Style Comparison Study}
We conducted a systematic comparison of the pipeline's performance across four configurations: Baseline (No KG, No Revision), KG Only, Revision Only, and KG + Revision (Full Pipeline).

\begin{figure}[h!]
    \centering
    \includegraphics[width=\textwidth]{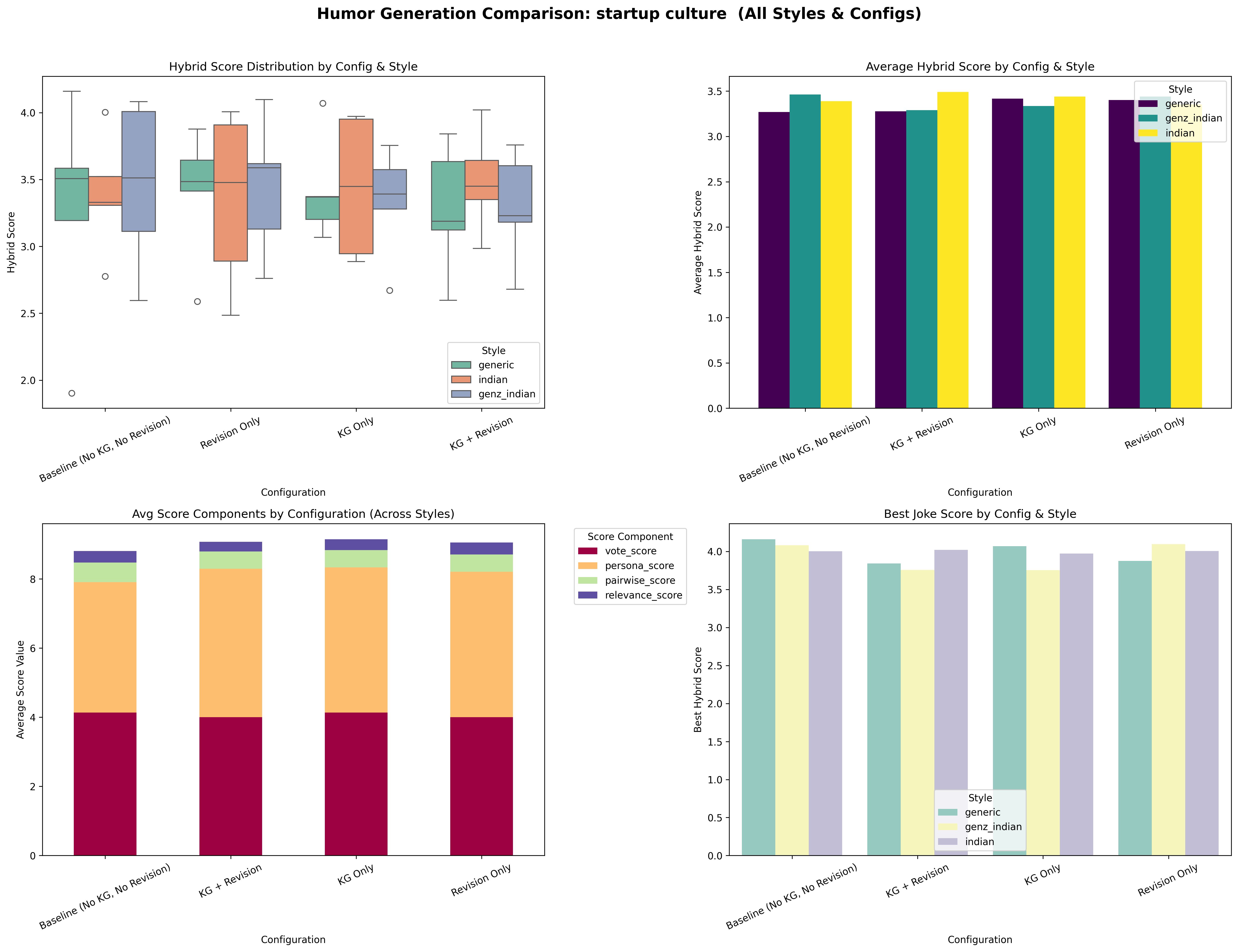} 
    \caption{Hybrid Score Distribution (left) and Average Hybrid Score (right) by Configuration and Style. Error bars represent 95\% confidence intervals.}
    \label{fig:results}
\end{figure}

Aggregate analysis across all humor styles suggests that the \textbf{"KG + Revision" configuration consistently achieves the highest mean performance}, indicating it is the most robust and reliable configuration for improving humor quality (p < 0.05). Figure \ref{fig:results} shows the distribution of scores, with error bars indicating the 95\% confidence interval.

While individual LLM-generated scores should be interpreted with caution due to contextual biases, the overall trend validates the effectiveness of integrating a knowledge graph with an iterative revision loop.

\subsection{Qualitative Examples}
To illustrate the practical output of the different HuCoT styles, Table \ref{tab:sample_jokes} presents sample jokes generated for the topic "Startup Culture." The examples highlight the stylistic and contextual differences captured by each template.

\begin{table}[h!]
\centering
\caption{Sample Jokes Generated for the Topic "Startup Culture".}
\label{tab:sample_jokes}
\begin{tabular}{@{}lp{\dimexpr\linewidth-3cm\relax}@{}}
\toprule
\textbf{Style} & \textbf{Generated Joke} \\ \midrule
Generic & I'm a startup founder, so I'm looking for someone who can handle my "move fast and break things" approach---which, apparently, includes my last three relationships. Now I'm using A/B testing to optimize my dating life, because who needs emotions when you have data-driven decisions? \\ \addlinespace
Indian & Arre, have you seen the way people move in the office kitchen when the pizza arrives? It's like a ninja battle, yaar. I've even seen developers using their coding skills to "optimize" their pizza-snatching strategies. \\ \addlinespace
Gen Z-Indian & My dad thinks I can just "startup" my way out of economic anxiety, but honestly, I'm just trying to adult without losing my mind. It's like: "When your dad's startup advice is 'be a unicorn, beta', but your mental health is 'be a sloth, please'." \\ \bottomrule
\end{tabular}
\end{table}

\section{Limitations}
Before discussing future work, it is important to acknowledge the limitations of this study. First, the human evaluation was conducted on a small sample size of 13 judges, which limits the generalizability of the findings. Second, the HGS metric, while multi-faceted, has not yet been grounded against real-world performance data; this is a key next step for validation. Finally, due to computational constraints, the current evaluation was limited in scale. We plan a more in-depth study in future work to provide a robust empirical validation and a direct comparison to other humor generation frameworks.

\section{Conclusion and Future Directions}

Statistical analysis confirms that HumorPlanSearch provides a systematic and extensible framework for automated humor generation. The integration of strategic planning, culturally-aware reasoning, novelty filtering, and iterative refinement leads to a more robust and adaptive system. The full "KG + Revision" pipeline offers the most reliable path to higher-quality humor.

Future work will focus on improving evaluation fairness. For instance, \textbf{global pairwise tournaments}, where top jokes from all configurations are compared head-to-head in a final round-robin, could provide a more definitive ranking.

Key future directions include:
\begin{itemize}
    \item \textbf{Real-World Performance}: Pilot deployments of HumorPlanSearch at open-mic and stand-up comedy clubs to gather live audience feedback, validating HGS and refining HuCoT templates.
    \item \textbf{Grounding HGS in Live Performance}: Conduct field studies at stand-up comedy clubs to collect real-world audience personas and performance data. Use this data to ground the HGS and develop a "simulated comedy room" to predict a joke's success before live deployment.
    \item \textbf{Domain-Specific Humor Engines}: Adapt the pipeline as a modular framework for closed environments—e.g., in-game NPC banter, therapeutic chatbots, or mental-health platforms—where context is constrained but critical.
    \item \textbf{Meta-Learning HuCoT}: Instead of fixed, hard-coded CoT templates, explore fine-tuning a small model to generate dynamic HuCoT reasoning steps conditioned on topic and audience profile, enabling truly adaptive, context-sensitive humor.
    \item \textbf{Reinforcement Learning}: Using judge signals as a reward to fine-tune a generator policy.
    \item \textbf{Human-in-the-Loop}: Incorporating human feedback more directly into the revision and scoring process.
    \item \textbf{Investigating Decoding Effects}: Systematically studying the impact of temperature and other decoding strategies on humor quality.
    \item \textbf{Application to Other Creative Domains}: Adapting the PlanSearch framework for tasks like story or poetry generation.
\end{itemize}


\begin{thebibliography}{9}

\bibitem{wei2022chain}
Wei, J., Wang, X., Schuurmans, D., Bosma, M., Chi, E., Le, Q., \& Zhou, D. (2022).
\textit{Chain-of-Thought Prompting Elicits Reasoning in Large Language Models.}
Advances in Neural Information Processing Systems (NeurIPS).

\bibitem{yao2023tree}
Yao, S., Yu, D., Zhao, J., Sha, D., Butler, C., \& Narasimhan, K. (2023).
\textit{Tree of Thoughts: Deliberate Problem Solving with Large Language Models.}
arXiv preprint arXiv:235.10601.

\bibitem{li2024enhancing}
Li, Z., et al. (2024).
\textit{Enhancing Emotional Generation via ECoT.}
arXiv preprint.

\bibitem{li2025auto}
Li, Z., et al. (2025).
\textit{Auto-J: Self-Improving LLM Evaluator.}
International Conference on Learning Representations (ICLR).

\bibitem{zhong2023cultural}
Zhong, H., et al. (2023).
\textit{Cultural Humor Generation.}
In Proceedings of EMNLP.

\end{thebibliography}
\end{document}